\documentclass[letterpaper, 10 pt, conference]{ieeeconf}  

\IEEEoverridecommandlockouts                              

\usepackage{amsmath,amsfonts}
\usepackage{algorithmic}
\usepackage{algorithm}
\usepackage{array}
\usepackage{textcomp}
\usepackage{stfloats}
\usepackage{url}
\usepackage{verbatim}
\usepackage{graphicx}
\usepackage{cite}

\usepackage[margins]{trackchanges}
\addeditor{SC}

\usepackage{graphics} 
\usepackage{svg}
\usepackage[table,x11names]{xcolor}
\usepackage[acronyms]{glossaries}
\usepackage{booktabs} 
\usepackage{siunitx} 
\usepackage{makecell} 
\usepackage{tabularx}
\usepackage{listings}
\usepackage{amssymb}

\let\labelindent\relax
\usepackage{enumitem} 
\usepackage{color}
\usepackage{xparse}
\usepackage{ifthen}
\usepackage[normalem]{ulem}

\usepackage{pifont}  
\usepackage{caption}    
\usepackage{subcaption} 
\usepackage{multirow} 
\usepackage{longtable}
\usepackage{multicol} 
\usepackage{diagbox} 
\usepackage{arydshln}
\usepackage{nicematrix}
\usepackage{tikz}
\usetikzlibrary{decorations.markings, patterns,shapes.arrows, arrows.meta, positioning, shapes}
\usepackage{pgfmath}
\usepackage{pgfplots}
\usepackage{pgfplotstable}
\usepackage{pgfplots}
\pgfplotsset{compat=newest}   
\usepgfplotslibrary{groupplots}
\usetikzlibrary{calc}
\usepackage{fontawesome5}
\usepgfplotslibrary{polar,fillbetween}
\allowdisplaybreaks
\usepackage{amsmath,amssymb,bm}
\usepackage{cuted}
\usepackage{mathtools}
\makeatletter
\let\NAT@parse\undefined
\makeatother
\usepackage[hidelinks]{hyperref}
\usepackage{cleveref} 

\DeclareMathSizes{10}{9}{7}{5}

\allowdisplaybreaks
\newtheorem{definition}{Definition}

\makeatletter
\def\CheckMinus\ignorespaces{\@ifnextchar-{}{\phantom{-}}}
\makeatother
\newcolumntype{I}{>{\CheckMinus}c}
\newcolumntype{J}{>{\CheckMinus}l}
\newcolumntype{H}{>{\setbox0=\hbox\bgroup}c<{\egroup}@{}}
\newcommand{\mathscale}{0.8}

\newcommand{\Tvec}{T}
\newcommand{\tc}{t}

\newcommand{\Fmat}{\mathbf{F}}

\newcommand{\Gmat}{\mathbf{G}}
\newcommand{\uref}{u^{\text{r}}}

\newcommand{\PDij}{P_{D}(i,j)}

\def \ij {{i,j}}

\newcommand{\Xs}{\mathcal{D}}

\newcommand{\Xu}{\mathcal{D}^{\neg}}
\newcommand{\Xsc}{\mathcal{P}}
\newcommand{\Fcal}{\mathcal{F}}
\newcommand{\eye}{E}

\newcommand{\Eb}{\mathbf{\eye}}

\newcommand{\Cmat}{\mathbf{C}}
\newcommand{\Ccal}{\mathcal{C}}

\newcommand{\Scal}{\mathcal{S}}

\newcommand{\uvec}{\mathbf{u}}
\newcommand{\pvec}{\mathbf{p}}

\newcommand{\Qcal}{\mathcal{Q}}
\newcommand{\Acal}{\mathcal{A}}
\newcommand{\zl}{\underline{z}}
\newcommand{\zu}{\overline{z}}
\newcommand{\Qucal}{\mathcal{Q}_u}
\newcommand{\Qxcal}{\mathcal{Q}_x}
\newcommand{\Qpcal}{\mathcal{Q}_p}
\newcommand{\bM}{Big-M }

\newcommand\zero{{\boldsymbol{0}}}
\newcommand{\ones}{\mathbf{1}}

\def \nu {n_u}

\newcommand{\xvec}{\mathbf{x}}

\newcommand{\Qu}{Q_{u}}

\newcommand{\blkdiag}{\scalebox{1.5}{$\Phi$}}

\newcommand{\cvec}{\mathbf{c}}
\newcommand{\pub}[1]{\overline{\pvec}_{#1}}
\newcommand{\plb}[1]{\underline{\pvec}_{#1}}

\newcommand{\iden}{\mathbf{I}}

\newcommand{\Hnull}{$\text{H}_0$}

\newcommand{\Hnulleq}{$\text{H}_0^{=}$}
\newcommand{\Hnullgeq}{$\text{H}_0^{\geq}$}
\newcommand{\Hnullleq}{$\text{H}_0^{\leq}$}

\newcommand{\pone}{$0.05^{*}$}
\newcommand{\ptwo}{$0.01^{**}$}
\newcommand{\pthree}{$0.001^{***}$}

\newcommand{\layoutlabel}[4]{%
    \draw[white, line width=2pt, >=Latex] (#1) -- (#2);
    \node[
      circle,
      line width=0pt,
      draw=white,
      scale=0.8,
      fill=white,
      font=\normalsize\bfseries,
    ] at (#2) {#3};
}

\newcommand{\layoutlegend}[4]{%
\node[
  scale = 0.9,
  circle,
  draw=black,
  line width=1.pt,
  fill=white,
  font=\footnotesize\bfseries,
  inner sep=1pt
] at (#1) {#3};
\node[anchor=west, scale = 1, font=\footnotesize, align=left, text=black] at (#2) {#4};
}

\newcommand\hyp[1]{\textbf{H{#1}}}
\newcommand{\ms}{significant }

\newcommand{\vhs}{very highly significant }

\def\eg{\textit{e.g.},\ }

\newcommand{\figref}[1]{\hyperref[#1]{Fig.~\ref*{#1}}}
\newcommand{\tabref}[1]{\hyperref[#1]{Tab.~\ref*{#1}}}
\newcommand{\secref}[1]{\hyperref[#1]{Sec.~\ref*{#1}}}

\newcommand{\figvspace}{\vspace{-0.4cm}}
\newacronym{lfd}{LfD}{Learning from Demonstration}
\newacronym{mpc}{MPC}{Model Predictive Control}
\newacronym{mimpc}{MI-MPC}{Mixed-Integer Model Predictive Controller}
\newacronym{miqp}{MIQP}{Mixed-Integer Quadratic Programming}
\newacronym{mip}{MIP}{Mixed-Integer Programming}
\newacronym{qp}{QP}{Quadratic Program}
\newacronym{st}{s.t.}{subject to:}
\newacronym{mcis}{MCIS}{Maximal Control Invariant Set}
\newacronym{cis}{RCIS}{Robust Controlled Invariant Set}
\newacronym{cbf}{CBF}{Control Barrier Function}
\newacronym{dtls}{DTLS}{Discrete-Time Linear System}
\newacronym{cocp}{COCP}{Constrained Optimal Control Problem}
 
\newacronym{tlx}{Raw-TLX}{NASA Raw Task Load Index}
\newacronym{A}{$\mathbf{CAAC}$}{Constraint-Aware Assistive Controller}
\newacronym{NA}{$\mathbf{UOC}$}{User-Only Control}
\newacronym{RTQ}{RTQs}{RAW-TLX Questions}
\newacronym{AQ}{AQs}{Additional Questions}
\newacronym{OEQ}{OEQs}{Open-ended Questions}
\newacronym{WSRT}{WSR}{Wilcoxon Signed-Rank}
\newacronym{MWT}{MWU}{Mann-Whitney U}
\newacronym{SWT}{SW}{Shapiro-Wilk}

\newacronym{md}{\textbf{MD}}{\text{Mental Demand}}
\newacronym{pd}{\textbf{PD}}{\text{Physical Demand}}
\newacronym{td}{\textbf{TD}}{\text{Temporal Demand}}
\newacronym{sc}{\textbf{SC}}{\text{Successful}}
\newacronym{eff}{\textbf{EF}}{\text{Effort}}
\newacronym{fr}{\textbf{FR}}{\text{Frustration}}
\newacronym{sf}{\textbf{SF}}{\text{Safe}}
\newacronym{ct}{\textbf{CT}}{\text{Control}}
\newacronym{cd}{\textbf{CD}}{Completion Duration}
\newacronym{nc}{\textbf{NC}}{\text{Number of Collisions}}

\title{\LARGE \bf
Minimal Intervention Shared Control with Guaranteed Safety under Non-Convex Constraints
}

\author{Shivam Chaubey, Francesco Verdoja, Shankar Deka, Ville Kyrki%
\thanks{The authors acknowledge the use of the MIDAS infrastructure of Aalto School of Electrical Engineering.
S.\ Chaubey, F.\ Verdoja, S.\ Deka, and V.\ Kyrki are with the School of Electrical 
Engineering, Aalto University, Espoo, Finland. 
{\tt\small \{firstname.lastname\}@aalto.fi}}%
}

\begin{document}

\maketitle

\begin{abstract} 
Shared control combines human intention with autonomous decision-making. At the low level, the primary goal is to maintain safety regardless of the user’s input to the system. However, existing shared control methods—based on, e.g., \glsentrylong{mpc}, \glsentrylongpl{cbf}, or learning-based control—often face challenges with feasibility, scalability, and mixed constraints.

To address these challenges, we propose a \glsentrylong{A} that computes control actions online while ensuring recursive feasibility, strict constraint satisfaction, and minimal deviation from the user’s intent. It also accommodates a structured class of non-convex constraints common in real-world settings. We leverage \glsentrylongpl{cis} for recursive feasibility and a \glsentrylong{miqp} formulation to handle non-convex constraints.
We validate the approach through a large-scale user study with 66 participants—one of the most extensive in shared control research—using a simulated environment to assess task load, trust, and perceived control, in addition to performance. The results show consistent improvements across all these aspects without compromising safety and user intent. Additionally, a real-world experiment on a robotic manipulator demonstrates the framework’s applicability under bounded disturbances, ensuring safety and collision-free operation.
\end{abstract}

\section{Introduction}\label{sec:intro}

Shared control is a framework where control authority is either continuously or intermittently shared between a human operator and an assistive system~\cite{doi-10-1177-0278364913490324,doi-10-1115-1-4039145}, and is commonly used in teleoperation, semi-autonomous navigation, assistive robotics, and medical systems. It combines user intent with the enforcement of safety constraints, task-specific requirements, or optimal behavior. In practice, users often have implicit goals but limited knowledge of internal dynamics or system constraints (\eg joint limits, actuator capabilities, or environmental factors), which may lead to unsafe situations. Thus, assistive systems are needed to ensure safety and constraint satisfaction without compromising user control intent~\cite{10590767}. Although user intent prediction and task-driven shared control can enhance autonomy, they risk overriding user control, diminish trust, and depend on accurate demonstrations for model learning~\cite{javdani2015shared,doi-10-1177-0278364913490324}. Preserving intent is thus crucial for user engagement and trust, especially in human-centered applications~\cite{8643740}.

\begin{figure}[t]
  \centering
  \input{images/cover_img/cover_img}
  \caption{Shared control framework: minimal adjustment of user input to avoid unsafe regions (red).}
  \label{fig:game_layout}
  \figvspace
\end{figure}

\glsresetall{}
Equally important is strict constraint satisfaction, which arises from physical limitations (\eg actuator bounds or robot configurations), environmental considerations (\eg obstacle avoidance), and task-specific demands~\cite{10802533} (\eg precise manipulation in activities like pouring~\cite{kronhardt2023understandingsharedcontrolassistive}). When these constraints are not strictly met, safety, feasibility, task success, and user trust can be compromised. Methods such as \gls{mpc} and \glspl{cbf} can enforce constraints, but in practice often require tuning of horizon length or slack penalties~\cite{8643740,8651894,7317796,10202565,broad2019highlyparallelizeddatadrivenmpc} and barrier-function design (hand-crafted or learned)~\cite{10156043,robey2020learningcontrolbarrierfunctions,dawson2021safenonlinearcontrolusing}, and may be difficult to implement reliably under strict non-convex state constraints and bounded control inputs~\cite{9453404,srinivasan2020synthesiscontrolbarrierfunctions}. Learning-based assistance can be context-aware~\cite{losey2021learninglatentactionscontrol,10015627}, but typically lacks hard guarantees beyond training data. These limitations motivate a unified framework that preserves user intent while providing strict constraint satisfaction and recursive feasibility without extensive tuning in such settings.

To address these limitations, we propose a \gls{A} framework  
that ensures recursive feasibility and safety by leveraging \glspl{cis}, enabling online, constraint-aware assistance via a single-step optimization. It also addresses a structured class of non-convex safe sets--specifically, those representable as the complement of a finite union of convex polytopic unsafe sets. 

Our main contributions are as follows.
First, we develop our framework by integrating offline-computed \glspl{cis} into a single-step \gls{cocp}, reformulated as a \gls{miqp} to handle non-convex constraints.
Second, we validate its effectiveness against our hypotheses through a large-scale user study with 66 participants—the first of its kind at this scale in shared control research.
The study shows reduced workload, increased trust, preserved control authority, satisfied safety constraints, and improved performance.
Finally, we validate the method on a robotic manipulator, where the controller maintains feasibility under bounded disturbances and ensures collision-free operation\footnote{Supplementary material and code are available at: \url{https://version.aalto.fi/gitlab/irobotics/CAAC}}. 
\section{Related Work} \label{sec:related_work}
A wide range of shared control strategies have been proposed to assist users in task completion. These approaches differ in how they interpret user intent, ensure safety, and preserve the user's sense of control. 

Probabilistic methods, such as hindsight optimization~\cite{javdani2015shared} and policy blending~\cite{doi-10-1177-0278364913490324}, anticipate user intention and assist with task completion. Hindsight optimization learns the user’s goal through Partially Observable Markov Decision Processes (POMDPs). Policy blending, on the other hand, infers user intent and adjusts the controls accordingly. While both methods offer more proactive autonomy, they lack strict safety guarantees. Moreover, users often report a loss of control when the system enforces a specific inferred strategy, limiting their ability to adapt. This highlights a trade-off between autonomous assistance and user adaptability.

In the context of vehicles, shared control literature frequently focuses on predictive methods such as \gls{mpc} to enforce constraints and optimize behavior. Without a suitable terminal/invariant set, ensuring recursive feasibility can require careful horizon selection and constraint handling, and obstacle anticipation often motivates longer horizons that increase computational cost~\cite{electronics13081442}. 

To reduce this cost, some works adopt variable-step horizons~\cite{7317796,schwarting2017parallel}, and model obstacle avoidance via soft constraints~\cite{8643740,7317796} or cost penalties~\cite{8651894,10202565}. Other works emphasize user authority or reduce early intervention~\cite{8643740,schwarting2017parallel}. Overall, challenges remain in achieving hard safety guarantees in practice with limited tuning and real-time computation, and in adequately evaluating user experience.

\gls{cbf}-based methods provide theoretical safety via forward invariance safe sets, and are commonly implemented through real-time \gls{qp}/\gls{mpc} safety filters or potential-field style controllers. In practice, barrier functions are either hand-crafted~\cite{doi-10-1145-3594806-3596522} or learned from data~\cite{robey2020learningcontrolbarrierfunctions}, while feasibility is often enforced through slack variables~\cite{10590767,dallas2025controlbarrierfunctionsshared,7799015}. However, many \gls{cbf}-based frameworks do not directly accommodate strict non-convex state and control constraints~\cite{9453404,srinivasan2020synthesiscontrolbarrierfunctions}, and extensive user studies remain limited.

Learning-based methods can adapt robot impedance and provide haptic guidance based on task dynamics, as discussed in \cite{10272651}. Human behavior modeling using inverse differential games is explored in \cite{10056301}. Operator workload reduction by learning manipulation tasks from demonstrations is proposed in \cite{doi-10-1177-1729881419857428}. While these approaches are promising for specific tasks, they often lack safety guarantees and restrict the user to learned tasks, thereby limiting flexibility.

To address the need for strict safety guarantees, feasibility, and preservation of user control, we propose leveraging \gls{cis}, which ensures that, from any state within the set, there exists an admissible control input that keeps the system safe.
Recent works~\cite{fiacchini2017computingcontrolinvariantsets,doi-10-1145-3365365-3382205,10328804}—including implicit formulations, lifted-space methods, and efficient set approximations—have made \gls{cis} practical for high-dimensional systems. While \gls{cis} have been widely used in control theory, their application to shared control remains limited. Compared to approaches that often rely on learned or hand-designed barrier functions and feasibility tuning, \gls{cis} provide a systematic way to certify safety and recursive feasibility for admissible control actions within the set.

Building on these insights, we first formulate the \gls{cocp} problem for our \gls{A} framework, then embed \glspl{cis} to guarantee recursive feasibility, and finally handle non-convex constraints within a \gls{miqp} formulation.
\section{Problem Formulation}\label{sec:problem_formulation}

To formalize the shared control problem, we consider a \gls{dtls} of the form: 
\begin{equation} x_{k+1} = A x_k + B u_k + E w_k, \label{eq:dtls} \end{equation} 
where $x_k \in \mathbb{R}^n$ denotes the state vector, $u_k \in \mathbb{R}^m$ the control input, and $w_k \in \mathbb{R}^d$ an unknown disturbance at time step $k$. The matrices $A \in \mathbb{R}^{n \times n}$, $B \in \mathbb{R}^{n \times m}$, and $E \in \mathbb{R}^{n \times d}$ define the system dynamics. The disturbance $w_k$ is assumed to lie within a bounded set $\mathcal{W} = \{ w \in \mathbb{R}^d \mid G_w w \le g_w \},$ where $G_w \in \mathbb{R}^{p \times d}$ and $g_w \in \mathbb{R}^p$ define the polyhedron.

\paragraph{State Constraints} \label{sec:sate_cons}

We define the feasible region for states as a convex set $\Xsc$ and a non-convex set $\Xs$.

The convex set is represented as a polytope:
\begin{equation} 
\Xsc = \left\{ x \in \mathbb{R}^{n} \mid \Fmat x \leq f \right\}, 
\label{eq:state_inc_cons} \end{equation} 
where $\Fmat \in \mathbb{R}^{m_F \times n}$ and $f \in \mathbb{R}^{m_F}$. 
\begin{figure}
\centering
\begin{subfigure}[t]{0.22\textwidth}
\centering
\begin{tikzpicture}[scale=0.75]
\node at (4.5,3.9) {\large $\mathbb{R}^{n}$};

\draw[thick] (0.0,0.0) rectangle (4.6,3.6);
\node at (0.3,3.3) {\large $\mathcal{P}$};

\coordinate (A) at (2,2.5);
\coordinate (B) at (2.5,3.5);
\coordinate (C) at (3,2.5);
\draw[fill=gray!40] (A) -- (B) -- (C) -- cycle;

\path let 
  \p1 = ($ (A) + (B) + (C) $),
  \n1 = {veclen(\x1,\y1)/3},
  \n2 = {\x1/3},
  \n3 = {\y1/3}
in 
  node at (\n2,\n3) {$\Xu_1$};

\coordinate (R1) at (1,1);
\coordinate (R2) at (1.8,1.8);
\draw[fill=gray!40] (R1) rectangle (R2);
\path let 
  \p2 = ($0.5*(R1) + 0.5*(R2)$)
in 
  node at (\p2) {$\Xu_2$};

\coordinate (D1) at (3.5,1);
\coordinate (D2) at (4,1.5);
\coordinate (D3) at (4.5,1);
\coordinate (D4) at (4,0.5);
\draw[fill=gray!40] (D1) -- (D2) -- (D3) -- (D4) -- cycle;

\path let 
  \p3 = ($ (D1) + (D2) + (D3) + (D4) $),
  \n4 = {\x3/4},
  \n5 = {\y3/4}
in 
  node at (\n4,\n5) {$\Xu_3$};

\node at (2.3,-0.3) {$\Xs = \mathbb{R}^{n} \setminus (\Xu_1 \cup \Xu_2 \cup \Xu_3)$};
\end{tikzpicture}
\vspace{-1.7em}
\caption{}
\label{fig:nonconvex_safe_sets_a}
\end{subfigure}
\hspace{0.5em}
\begin{subfigure}[t]{0.22\textwidth}
\centering
\begin{tikzpicture}[scale=0.75]
\node at (4.5,3.9) {\large $\mathbb{R}^{n}$};

\draw[thick] (0.0,0.0) rectangle (4.6,3.6);
\node at (0.3,3.2) {\large $\mathcal{P}$};

\coordinate (T1) at (3.2,0.6);
\coordinate (T2) at (3.5,1.6);
\coordinate (T3) at (4.2,0.7);
\draw[fill=gray!30] (T1) -- (T2) -- (T3) -- cycle;
\path let \p3 = ($ (T1) + (T2) + (T3) $),
          \n5 = {\x3/3},
          \n6 = {\y3/3}
in node at (\n5,\n6) {$\Xu_3$};

\coordinate (P1) at (0.5,0.5);
\coordinate (P2) at (2,2.5);
\coordinate (P3) at (4,2);
\coordinate (P4) at (1.8,1);
\coordinate (P5) at (2.5,0.5);

\draw[fill=gray!30] (P1) -- (P2) -- (P3) -- (P4) -- (P5) -- cycle;

\path let 
  \p1 = ($(P2)-(P1)$),   
  \p2 = ($(P4)-(P1)$),   
  \n1 = {(\x1*\x2 + \y1*\y2)/(\x1*\x1 + \y1*\y1)} 
in
  coordinate (Int) at ($ (P1) + \n1*(\p1) $);

\path let 
  \p5 = ($ (P1) + (Int) + (P5) $),
  \n7 = {\x5/3},
  \n8 = {\y5/3}
in
  node at (\n7,\n8) { $\Xu_1$};

\path let 
  \p6 = ($ (P2) + (P3) + (P4) + (Int) $),
  \n9 = {\x6/4},
  \n{10} = {\y6/4}
in
  node at (\n9,\n{10}) { $\Xu_2$};

\path let 
  \p1 = ($(P2)-(P1)$),
  \n1 = {atan2(\y1,\x1)}
in
  node[rotate=\n1] at (1.10,1.75) { $\Xu_{1,2} = \Xu_1 \cup \Xu_2$};

\draw[dashed] (Int) -- (P4);
\node at (2.3,-0.3) {$\Xs = \mathbb{R}^{n} \setminus (\Xu_1 \cup \Xu_2 \cup \Xu_3)$};
\end{tikzpicture}
\vspace{-1.7em}
\caption{}
\label{fig:nonconvex_safe_sets_b}
\end{subfigure}
\caption{Illustration of (a) convex unsafe sets and (b) a non-convex unsafe region decomposed into convex components.}
\label{fig:nonconvex_safe_sets}
\figvspace{}
\end{figure}

In this work, we define the non-convex safe set $\Xs$ as the complement of an unsafe set: $\Xs = \mathbb{R}^n \setminus \Xu$. The unsafe set $\Xu$ is modeled as a finite union of convex polytopes: $\Xu = \bigcup_{i=1}^{n_b} \Xu_i$, where each $\Xu_i$ is either a convex region in itself (\figref{fig:nonconvex_safe_sets_a}) or a component obtained by decomposing a non-convex unsafe region into convex parts (\figref{fig:nonconvex_safe_sets_b}). The set $\mathcal{X} = \Xsc \cap \Xs$ defines the state admissible region in which the state $x$ must remain at each time step.

\paragraph{Control constraints} 
Control constraints ensure that the control input $u$ lies within a feasible region represented as a convex polytope:
\begin{equation}        
    \mathcal{U} = \{ u \in \mathbb{R}^{m} \mid \Gmat u \leq g \}, \label{eq:cont_cons}
\end{equation}
where $\Gmat \in \mathbb{R}^{m_G \times m}$ and $g \in \mathbb{R}^{m_G}$ define the half-space representation (H-representation) of the polytope.

\paragraph{Assistive Control Problem}
The objective of \gls{A} is to determine a control input $u_k$ that minimally perturbs the user's input $\uref_k$ when necessary to ensure that safety and task-specific constraints are satisfied for all future time steps. Although the optimization considers only the current input, constraints are enforced over an infinite horizon to ensure that the selected control does not lead the system into future states where constraint satisfaction becomes infeasible. This is particularly important in shared control, where user inputs may inadvertently drive the system toward the boundary of the infeasible region, requiring early intervention to avoid entering infeasible or unsafe states. Given $x_{k+h}  \in  \Xsc \cap \mathcal{D}$, this problem can be formulated as:
\begin{subequations}
\begin{align}
     &J_u = \min_{u_k}  \frac{1}{2} \big(u_{k} - \uref_k \big)^\top \Qu \big(u_{k} - \uref_k \big) \label{eq:prim_obj_obj} \\
&
\begin{aligned}
& {\text{s.t.}\ \ } \exists u_{k+h} \in \mathcal{U}, \forall h \geq 0,\ \ {\text{for which; }\ } \\
     & { Ax_{k+h} + Bu_{k+h} + E w_{k+h} \in  \Xsc \cap \mathcal{D}},\  \forall w_{k+h} \in \mathcal{W}, 
\end{aligned}     
     \label{eq:prim_obj_cons} 
\end{align}
\label{eq:prim_obj}
\end{subequations}
where $\Qu$ is a positive definite weight matrix.
\section{Methods}\label{sec:methods}

The problem \eqref{eq:prim_obj} enforces infinite-horizon constraints, solving it directly is intractable. We instead constrain the system to a precomputed \gls{cis}, guaranteeing recursive feasibility and reducing the problem to a single-step optimization solvable in real time.

To address non-convex state constraints, the safe region—defined as the complement of a union of convex polytopic unsafe sets—is encoded using the \bM method~\cite{schouwenaars2001mixed}, which converts the disjunctive structure into an equivalent intersection of relaxed half-spaces. The resulting constraints are constructed in sparse form within the \gls{miqp} framework, ensuring compatibility with solver and efficient use of sparsity. Together with one-step optimization and the use of \gls{cis}, this formulation enables real-time shared control. While \gls{miqp} problems have exponential worst-case complexity, the combination of one-step structure and sparsity greatly reduces solver overhead in practice.

\paragraph*{Notation}

Let $\blkdiag(A_1, A_2, \dots, A_n)$ denote a block-diagonal matrix with blocks $A_i \in \mathbb{R}^{a \times b}$. The vectors $\ones_a$ and $\zero_a$ are column vectors of ones and zeros of size $a \times 1$, and $\zero_{a,b}$ is the $a \times b$ zero matrix. Time indices are omitted when clear from context: the state $\mathbf{x}_k = [x_k, x_{k+1}]$, control $u_k$, and binary variable $p_{i,j,k+1}$ are written simply as $\mathbf{x}$, $\mathbf{u}$, and $p_{i,j}$.
\subsection{Robust Controlled Invariant Sets}
\begin{definition}
\label{def:mcis}
Consider the discrete-time linear system~\eqref{eq:dtls}, subject to state constraints $x_k \in \mathcal{X} \subseteq \mathbb{R}^n$ and control constraints $u_k \in \mathcal{U} \subseteq \mathbb{R}^m$, where $\mathcal{X} = \Xsc \cap \Xs$. Let $\mathsf{S}_{x,u} = \mathcal{X} \times \mathcal{U}$ denote the set of admissible state-control pairs.

A set $\mathcal{S} \subseteq \mathcal{X}$ is called a \gls{cis} if, for every $x_k \in \mathcal{S}$, there exists a control input $u_k \in \mathcal{U}$ such that the next state $x_{k+1} = A x_k + B u_k + E w_k$ also belongs to $\mathcal{S}$ for all $w_k \in \mathcal{W}$. 
\end{definition}

\par
Equivalently, if the system starts at any $x_k \in \mathcal{S}$, there exists a control sequence $\{u_k\}_{k \geq 0} \in \mathcal{U}$ that keeps the system within $\mathcal{S} \subseteq \mathcal{X}$ under disturbance $w_k$, ensuring recursive feasibility and constraint satisfaction at all future time steps. In this work, $\mathcal{S}$ denotes a polytopic representation of the \gls{cis} for the admissible set $\mathsf{S}_{x,u}$, computed offline using the explicit closed-form method proposed in~\cite{10328804}.

\subsection{\textbf{COCP} formulation}
To ensure recursive feasibility while minimizing deviation from the user's control input, a safe \gls{cocp} is formulated by explicitly incorporating the \gls{cis} as a terminal constraint, $x_{k+1} \in \mathcal{S}$. Given $x_k \in \mathcal{S}$, the resulting optimization problem at $k^{\text{th}}$ time-step is:
\begin{equation}
\begin{split}
J_u &= \min_{u} \frac{1}{2} \big(u_{k} - \uref_k\big)^\top \Qu \big(u_{k} - \uref_k\big) \\
 \text{s.t.} \quad &x_{k+1} = A x_{k} + B u_{k} + E w_{k}, u_k \in \mathcal{U}, \\
& x_{k+1} \in \mathcal{S} \subseteq \mathcal{X}, \forall w_k \in \mathcal{W}.
\label{eq:opt_cis}
\end{split}
\end{equation}

\subsection{Quadratic Programming Formulation}

Reformulating the problem in \eqref{eq:opt_cis} as a standard \gls{miqp} allows us to encode non-convex constraints, while structuring it in sparse matrix form provides a solver-ready formulation that can exploit sparsity for online \gls{A} applications.

The \gls{cocp} problem \eqref{eq:opt_cis} can thus be reformulated as:
\begin{subequations}
\label{eq:MPC_QP}
\begin{align}
    J &= \min \ \frac{1}{2} z^\top \Qcal z + q^\top z \label{eq:obj} \\
    \text{s.t.}& \quad \zl \leq \Acal z \leq \zu \label{eq:cons} .
\end{align}
\end{subequations}
Here, $z$ is the decision vector including states, controls, and binary variables introduced to handle non-convex constraints. $\Qcal$ is the positive semi-definite Hessian matrix, and $q$ is the corresponding linear cost vector. The matrix $\Acal$ encodes all equality and inequality constraints, while $\zl$ and $\zu$ denote the element-wise lower and upper bounds, respectively. 

\subsubsection{Equality State Constraints}
Since the optimization is performed over a single step, we define the stacked state and control vectors as $\xvec = [x_{k}^\top,\ x_{k+1}^\top]^\top$ and $\uvec = u_{k}$ to maintain a consistent vector notation. 

The combined nominal \gls{dtls}~\eqref{eq:dtls} and initial condition can be written compactly as:
\begin{small}
\begin{equation}
\begin{bmatrix}
\mathcal{A}' & \mathcal{B}'
\end{bmatrix}
\begin{bmatrix}
\xvec\\
\uvec
\end{bmatrix}
= -\xvec_0,
\label{eq:sys_dyn_mat}
\end{equation}
\end{small}

\noindent $\xvec_0 = [x_k^\top, \zero_{n}^{\top}]^\top$ encodes the known initial state $x_k \in \Scal$ at time step $k$, and matrices $\mathcal{A}' \in \mathbb{R}^{2n \times 2n}$ and $\mathcal{B}' \in \mathbb{R}^{2n \times m}$ are defined as:
\begin{small}
\begin{equation*}
\mathcal{A}' = 
\begin{bmatrix}
\zero_{n , n} & \zero_{n , n} \\
A & \zero_{n , n}
\end{bmatrix} - \iden_{2n \times 2n},
\quad
\mathcal{B}' = 
\begin{bmatrix}
\zero_{n , m} \\
B
\end{bmatrix}.
\end{equation*}
\end{small}

\subsubsection{Convex State Inequality \gls{cis} Constraints}
For the admissible set of state--input pairs~\eqref{eq:state_inc_cons}--\eqref{eq:cont_cons} with $\mathsf{S}_{x,u} = \mathcal{P} \times \mathcal{U}$ and the \gls{dtls}~\eqref{eq:dtls}, we compute a \gls{cis}. The \gls{cis} is represented by the polytope  $\mathcal{S}_{F} = \left\{\, x \in \mathbb{R}^{n} \;\middle|\; \Cmat_{F} x \leq c_{F} \,\right\}.
$  To impose these constraints only on the successor state $x_{k+1}$, we use the modified representation:  
\begin{equation}
    \Cmat_{\Fcal} \xvec \leq c_{\Fcal}, \label{eq:state_con_mat}
\end{equation}
where  $
    \Cmat_{\Fcal} = [ \zero_{m_{\Cmat_F} \times n} \ \ \Cmat_F ],  
    \quad c_{\Fcal} = c_F,
$  
and $m_{\Cmat_F}$ denotes the number of rows of $\Cmat_F$.
\subsubsection{Non-Convex State Inequality \gls{cis} Constraints}
\paragraph{Decomposition of Non-Convex Safe Sets}
The convex unsafe region $\Xu_i$ is described as intersections of $n_{c_i}$ half-spaces, with each half-space $\PDij$ defined by a linear inequality:
\begin{small}
$\Tvec_{\ij}^\top x > \tc_{\ij},$
\end{small}
with $\Tvec_{\ij}$ is the inward-pointing normal vector of the $j^{\text{th}}$ half-space, and $\tc_{\ij}$ is the corresponding offset.
Thus, the region $\Xu_i$ is
\begin{small}
$\Xu_i = \cap_{j=1}^{n_{c_i}} \{ x \mid \Tvec_{\ij}^\top x > \tc_{\ij} \}.
$\end{small}
To define the safe region $\Xs_i$, we take the complement of the unsafe region $\Xu_i$, resulting in a union of half-spaces:
\begin{small}
$\Xs_i = \cup_{j=1}^{n_{c_i}} \{ x \mid \Tvec_{\ij}^\top x \leq \tc_{\ij} \}.$
\end{small}
Finally, the overall safe region $\Xs$ is defined as the intersection of the individual regions $\Xs_i$:
\begin{subequations} \label{eq:state_exc_cons}
\begin{equation}
\Xs = \left\{ x \in \mathbb{R}^n \;\middle|\; \cap_{i=1}^{n_b} \left( \cup_{j=1}^{n_{c_i}} \PDij \right) \right\},
\tag{\ref{eq:state_exc_cons}}
\end{equation}
\begin{equation}
\PDij = \left\{ x \in \mathbb{R}^n \;\middle|\; \Tvec_{\ij}^\top x \leq \tc_{\ij} \right\}.
\label{eq:pdij}
\end{equation}
\end{subequations}

\paragraph{Conversion of Non-Convex to Convex sets} \label{par:non_convex}
To convert each non-convex set $\Xs_i$, which is the union of convex half-space $\PDij$ region, into intersection of half-spaces $\PDij$, we apply \bM formulation using binary variables $p_{\ij} \in \{0, 1\}$. Each $p_{\ij}$ corresponds to a half-space $\PDij$ of the non-convex safe set $\Xs_{i}$. The binary variables determine whether the set $\PDij$ is active ($p_{\ij} = 0$) or relaxed ($p_{\ij} = 1$).
A large constant $M \gg 0$ is used to relax the constraint when $p_{\ij} = 1$, effectively deactivating it.
In this way, the union of convex regions is represented through the intersection of all constraints—active or relaxed—controlled by binary variables.

Thus, the corresponding safe-set can be defined as:
\begin{equation}
    \left\{ x \in \mathbb{R}^n \;\middle|\; \cap_{i=1}^{n_b} \left( \cap_{j=1}^{n_{c_i}} \left( \Tvec_{\ij}^{\top} x - t_{\ij} \leq M p_{\ij} \right) \right) \right\} .
    \label{eq:xs_MILP}
\end{equation}

To ensure that the system state $x$ lies within at least one convex region $\Tvec_{\ij}^{\top} x - t_{\ij} \leq 0$ of each non-convex safe set $\Xs_i$, we enforce: 
\begin{equation}
0 \leq \sum_{j=1}^{n_{c_i}} p_{\ij} \leq n_{c_i} - 1, \quad \forall i \in {1, \dots, n_b} .
\label{eq:big_m_xs_cons} 
\end{equation}
The above formulation of \eqref{eq:xs_MILP} and \eqref{eq:big_m_xs_cons}, applied for all \(i \in \{1, n_b\}\) and \(j \in \{1, n_{c_i}\}\), ensures that the user is not restricted to a specific region but can freely switch between regions that best satisfy the control objectives.

\paragraph{Imposing \gls{cis} Constraints}
For each convex region $\PDij$ that defines a half-space of a non-convex safe set $\Xs_i$, we define the corresponding set of admissible state-control pairs as: $ \mathsf{S}_{x,u}(i,j) = \left\{ (x, u) \in \mathbb{R}^n \times \mathbb{R}^m \ \middle| \Fmat x \leq f, \ T_{\ij} x \leq \tc_{\ij},\ \Gmat u \leq g \right\}$.

A \gls{cis}, $\mathcal{S}(i,j)$ is computed for each admissible set $\mathsf{S}_{x,u}(i,j)$. Each $\mathcal{S}(i,j)$ contains all states $x_k$ for which there exists a control input $u_k$ such that the successor state $x_{k+1} = A x_k + B u_k + E w_k$ remains within the set for all $w_k \in \mathcal{W}$, satisfying all constraints. These sets are represented as polytopes in the state space of the form $\mathcal{S}(i,j) = \{ x \in \mathbb{R}^n \mid \Cmat_{\ij} x \leq c_{\ij} \}$, where $\Cmat_{\ij} \in \mathbb{R}^{m_{C_{\ij}} \times n}$ and $c_{\ij} \in \mathbb{R}^{m_{C_{\ij}}}$ define the polytope.

Similar to~\eqref{eq:xs_MILP}, the enforcement of the \gls{cis} constraint at time step $k+1$ for any admissible set $\mathsf{S}_{x,u}(i,j)$ follows the same approach. The resulting safe set is defined as:

\begin{multline}
      \left\{ x_{k+1} \in \mathbb{R}^n \;\middle|\; \cap_{i=1}^{n_b} \left( \cap_{j=1}^{n_{c_i}} ( \Cmat_{\ij} x_{k+1} \right. \right. \\
      \left. \left. - c_{\ij} \leq M \mathbf{1}_{m_{C_{\ij}}} p_{\ij,k+1}  ) \right) \right\}.
    \label{eq:cis_MILP}
\end{multline}

Following the conversion of non-convex to union of convex constraint formulation, we use equation~\eqref{eq:cis_MILP} to impose the \gls{cis} inequality constraints corresponding to the $j^\text{th}$ subregion (half-space) of the non-convex safe set $\Xs_i$ at next time step $k+1$:
$
    \Cmat_{i,j} x_{k+1}
    - M 
    \ones_{m_{C_{i,j}}}
    p_{i,j,k+1}
    \leq 
    c_{i,j}
$
For the full stacked state vector $\xvec_k$, this constraint is written as:

\begin{equation*}    
\underbrace{[\zero_{m_{C_{i,j}} \times n},\, \Cmat_{i,j}]}_{\Ccal_{i,j}} \xvec_k  
- M 
 \ones_{m_{C_{i,j}}} p_{i,j,k+1}
\leq 
c_{i,j}.
\end{equation*}

For brevity, we omit the time indices and write $\mathbf{x}$ in place of $\mathbf{x}_k$ and $p_{i,j}$ in place of $p_{i,j,k+1}$.
To include all $\mathcal{S}_{i,j}$ for all $j \in \{1, \dots, n_{c_i}\}$ corresponding to the $i^\text{th}$ safe set $\Xs_i$, we write
 $ \Ccal_i \xvec - M L_{\pvec_{i}} \pvec_{i} \leq \cvec_i $, where $\Ccal_i = [\Ccal_{i,1}^\top, \cdots, \Ccal_{i,n_{c_i}}^\top ]^\top$, $L_{\pvec_{i}} = \blkdiag(\ones_{m_{C_{i,1}}}, \dots , \ones_{m_{C_{i, n_{c_i}}}})$, binary vector $\pvec_{i} = [p_{i,1}^\top, \dots, p_{i,n_{c_i}}^\top]^\top$, and $\cvec_i = [ c_{i,1}^\top, \cdots, c_{i,n_{c_i}}^\top ]^\top$.

Similarly, to impose the intersection over all sets $\Xs_i$ for $i \in \{1, \dots, n_b\}$, we obtain:
\begin{equation}
    \Ccal_{\mathcal{T}} \xvec - M \mathbf{L}_{\pvec} \pvec \leq \cvec_{\mathcal{T}},
    \label{eq:cis_mat}
\end{equation}
where $\Ccal_{\mathcal{T}} = [\Ccal_1^\top, \cdots, \Ccal_{n_b}^\top ]^\top$, $\mathbf{L}_{\pvec} = \blkdiag(L_{\pvec_{1}}, \dots, L_{\pvec_{n_b}})$ is a block-diagonal matrix of dimension $\left( \sum_{i=1}^{n_b} \sum_{j=1}^{n_{c_i}} m_{C_{i,j}} \right) \times n_p$, $\pvec = [\pvec_{1}^\top, \dots, \pvec_{n_b}^\top ]^\top$, and $\cvec_{\mathcal{T}} = [\cvec_1^\top, \cdots, \cvec_{n_b}^\top]^\top$.

\paragraph{Enforcing constraints}
From equation~\eqref{eq:big_m_xs_cons}, at least one region $\PDij$ belong to $\Xs_i$ must be active in order to ensure avoidance of $\Xu_i$ unsafe convex set for time step $k+1$.

This can be compactly written as: \par\vspace{-6pt}

\begin{small}
\begin{align*}
    \plb{i} &\leq 
    \ones^{\top}_{n_{c_i}}
    \underbrace{
    \begin{bmatrix}
    p_{i,1}^\top & p_{i,2}^\top & \cdots & p_{i, n_{c_i}}^\top
    \end{bmatrix}^{\top}
    }_{\pvec_{i}}
    \leq 
    \pub{i},
\end{align*}
\end{small}
\noindent
where $\plb{i} = \zero$, $\pub{i} = (n_{c_i} - 1)$. To enforce this constraint for all safe regions ($\forall i \in \{1, \dots, n_b\}$), we define:
\begin{equation}
\plb{} \leq \Eb \pvec \leq \pub{},  \label{eq:p_mat}
\end{equation}

\noindent
where $\Eb = \blkdiag( \ones^{\top}_{n_{c_1}}, \dots,  \ones^{\top}_{n_{c_{n_b}}})$, $\plb{} = \zero_{n_p}$, and $\pub{} = [(n_{c_1} - 1) \dots n_{c_{n_b}} - 1)]^{\top}$.

\subsubsection{\gls{qp} Constraint Matrices}
The decision variable is defined as $ z = [\xvec^\top\ \uvec^\top\ \pvec^\top]^\top $. Combining the control input constraints~\eqref{eq:cont_cons}, equality constraint~\eqref{eq:sys_dyn_mat}, convex~\eqref{eq:state_con_mat} and non-convex~\eqref{eq:cis_mat} \gls{cis} state inequality constraints, and binary variable constraints~\eqref{eq:p_mat}. The complete constraint set in~\eqref{eq:cons} can be expressed as $ \underline{\mathbf{z}} \leq \mathcal{A} \mathbf{z} \leq \overline{\mathbf{z}} $, where

\begin{small}
\begin{align}
\hspace{-0.14cm}
\zl &=
\begin{bmatrix}
-\xvec_0 \\ -\infty \\ -\infty \\ -\infty \\ \plb{}
\end{bmatrix}, 
\ \ 
\Acal =
\begin{bmatrix}
\mathcal{A}' & \mathcal{B}' & \zero \\
\Cmat_{\Fcal} & \zero & \zero \\
\Ccal_{\mathcal{T}} & \zero & -M \mathbf{L}_{\pvec} \\
\zero & \Gmat & \zero \\
\zero & \zero & \Eb
\end{bmatrix}, 
\ \
\zu =
\begin{bmatrix}
-\xvec_0 \\ c_{\Fcal} \\ \cvec_{\mathcal{T}} \\ g \\ \pub{}
\end{bmatrix}.
\label{eq:comb_mat}\end{align}
\end{small}

\subsubsection{\gls{qp} Objective Matrices}
 The total objective function is expressed as \( J = J_x + J_u + J_p \), corresponding to state, control, and binary variable costs, respectively. Since we only track the user-provided control input, $J_x = 0$ and $J_p = 0$. 

The objective reduces to the control cost:
$
J_u = \frac{1}{2} \uvec^\top \Qucal \uvec + q_u^\top \uvec.
\label{eq:obj_u}
$

The full objective function \eqref{eq:obj} is then:

\begin{small}
\begin{equation}
J = \min \frac{1}{2} z^\top \underbrace{
\blkdiag(
\Qxcal,  \Qucal,  \Qpcal)}_{\Qcal} z + \quad
\underbrace{\left[q_x^\top\ q_u^\top\ q_p^\top\right]}_{q^\top} z,
\label{eq:qp_mat}
\end{equation}
\end{small}

where $
\Qxcal = 0_{ 2n \times 2n}, \
\Qucal = \iden_{m \times m}, \
\Qpcal = 0_{n_p  \times n_p}, \ 
q_x = 0_{n \cdot 2}, \
q_u^{\top} = -{\uref}^{\top}\Qucal , \
q_p = 0_{n_p}
$, \text{and} $n_p = \sum_{i=1}^{n_b} n_{c_i}$ represents the total number of binary variables.
In this formulation, $ \Qcal $ is positive semi-definite, as only $ \Qucal $ contributes a strictly convex term. The worst-case computational complexity for this \gls{miqp} is $\mathcal{O}\left(2^{n_p} \cdot ( 2 n  + m)^3\right)$.

\section{User Study}\label{sec:experiments}
\glsreset{A}
\glsreset{NA}
This section presents a user study designed to evaluate the effectiveness of the proposed \gls{A}. Our goal is to investigate how its use impacts task load, trust, perceived control, and task completion (performance). The study involves two modes---one with \gls{A} and one with \gls{NA}, where the user task is to navigate a maze-like environment by guiding a robot toward goals while avoiding obstacles. Given concerns about user and machine safety arising from letting novice users interact without assistance on a real system, we decided to conduct the user study in a simulation setting. We first describe our hypotheses, simulation setup, and user study setup. We then explain the evaluation methodology and statistical testing approach used to assess our hypotheses. 

\subsection{Hypotheses}
The following hypotheses were formulated to evaluate the effectiveness of \gls{A} compared to \gls{NA}:
\begin{enumerate} 
    \item[\hyp{1:}] \gls{A} decreases task load from the user. \label{hyp:1}
    \item[\hyp{2:}] \gls{A} increases user’s trust of being able to complete the task safely and successfully. \label{hyp:2}
    \item[\hyp{3:}] \gls{A} does not decrease the user’s perception of being in control of the system. \label{hyp:3}
    \item[\hyp{4:}] \gls{A} decreases task completion time. \label{hyp:4}
\end{enumerate} 
\subsection{Experimental Setting}
To evaluate the hypotheses, we designed a simulation (shown in \figref{fig:game_layout}) in which the objective is to guide a robot, represented by a small blue circle (1), to sequentially reach green goal circles (2), with progress tracked by a goal-reached counter (3), while avoiding black obstacles (5). Navigation is restricted to safe gray regions (4). 

\subsubsection{Robot Behavior} The robot state is defined as $x_k = [X_k,\, Y_k,\, v^x_k,\, v^y_k]^\top$ and the control input as $u_k = [a^x_k,\, a^y_k]^\top$, where $(X_k, Y_k)$ denote position, $(v^x_k, v^y_k)$ velocity, and $(a^x_k, a^y_k)$ acceleration in $x$ and $y$ directions. All quantities are expressed in consistent, unitless values without mapping to a physical scale. The robot follows linear damped double-integrator dynamics $x_{k+1} = Ax_k + Bu_k$, where
\begin{equation}    
\scalebox{\mathscale}{$
\begin{aligned}
    A = 
    \begin{bmatrix}
    1 & 0 & \Delta t & 0 \\
    0 & 1 & 0 & \Delta t \\
    0 & 0 & 1 - \gamma \Delta t & 0 \\
    0 & 0 & 0 & 1 - \gamma \Delta t
    \end{bmatrix}, \quad
    B = 
    \begin{bmatrix}
    0.5 \Delta t^2 & 0 \\
    0 & 0.5 \Delta t^2 \\
    \Delta t & 0 \\
    0 & \Delta t
    \end{bmatrix}.
\end{aligned}
$}
\label{eq:agent_dyna}
\end{equation}

\begin{figure}
\centering


\begin{tikzpicture}[scale=0.06]
\fill[white] (1,7.0) rectangle (83.8,110.7);

\fill[black] (1,23.1) rectangle (83.8,110.7);

\def\btnsize{4}
\def\spacing{0.5}
\def\centerx{42.4}
\def\centery{104.1}

\foreach \i [count=\n from -1] in {1,2,3,4} {
    \pgfmathsetmacro\x{\centerx + (\n - 1.5) * (\btnsize + \spacing)}

    \draw[fill=gray!60, draw=gray!60, rounded corners=1pt]
        (\x,\centery) rectangle ++(\btnsize, \btnsize);

    \node[scale=0.7] at ({\x + \btnsize/2}, {\centery + \btnsize/2}) 
                {\textcolor{red}{\Large \faTimes}}; 

    \ifnum\i=4
        \layoutlabel{\x + \btnsize/2+2,\centery + \btnsize/2}{\x + \btnsize/2+9,\centery + \btnsize/2}{3}{1}
        
    \fi
}

\fill[gray!70] (5,27.1) rectangle (79.2,100.7);
\layoutlabel{25,60}{32,60}{4}{1}

\fill[black] (68.7,27.0) rectangle (79.7,47.7);
\fill[black] (12.7,33.1) rectangle (51.9,48.4);
\fill[black] (5.0,42.4) rectangle (62.5,48.4);
\fill[black] (5.0,76.0) rectangle (42.9,94.3);
\fill[black] (48.8,54.3) rectangle (73.6,100.7);
\layoutlabel{58,70}{66,70}{5}{1}

\fill[blue] (7.1,98.1) circle[radius=1.2];
\layoutlabel{8.3,98.1}{15.1,98.1}{1}{1}

\fill[green!70, opacity=1.0] (7.9,73.5) circle[radius=2.15];
\node[scale=1] at (7.9,73.5) {\textcolor{black}{\bfseries a}};
\layoutlabel{9.975,73.5}{16.9,73.5}{2}{1}

\fill[green!70, opacity=0.7] (75.9,97.1) circle[radius=2.15];
\node[scale=1] at (75.9,97.1) {\textcolor{black}{\bfseries b}};
\fill[green!70, opacity=0.5] (54.8,39.7) circle[radius=2.15];
\node[scale=1] at (54.8,39.7) {\textcolor{black}{\bfseries c}};
\fill[green!70, opacity=0.3] (9.5,38.9) circle[radius=2.15];
\node[scale=1] at (9.5,38.9) {\textcolor{black}{\bfseries d}};


\layoutlegend{87,100}{88,100}{1}{Robot}
\layoutlegend{87,90}{88,90}{2}{Current Goal}
\layoutlegend{87,80}{88,80}{3}{Goal Reached\\Counter}
\layoutlegend{87,70}{88,70}{4}{Safe Region}
\layoutlegend{87,60}{88,60}{5}{Unsafe Region}

\end{tikzpicture}
\vspace{-0.8cm}
\caption{Simulation layout with goal sequence (a~$\rightarrow$~d) shown in green for illustration; only one goal appears at a time.}
\label{fig:game_layout}
\figvspace{}
\end{figure}
The damping coefficient is set to $\gamma = 0.1$, causing velocity to build gradually with input and decay without input. Velocity and acceleration are bounded along both axes.

The user provides the reference acceleration $\uref_k$ through the joystick's right analog stick, where direction sets the acceleration vector and deflection magnitude controls its strength. 

The environment in~\figref{fig:game_layout} defines the robot's workspace bounded by outer walls and five axis-aligned rectangular obstacles (unsafe region). The constraints are expanded to account for the robot's shape, ensuring collision avoidance considers its geometry, not just its center point. The robot must stay within the environment bounds and avoid each obstacle by satisfying a union of axis-aligned half-space constraints.

Based on the system dynamics in \eqref{eq:agent_dyna} and the defined constraints, we constructed the admissible set $\mathsf{S}_{x,u}(i,j)$ for each $j^\text{th}$ half-space belonging to the $i^\text{th}$ obstacle. We then computed a \gls{cis} for every $\mathsf{S}_{x,u}(i,j)$, constructed the constraint matrices \eqref{eq:comb_mat} and objective \eqref{eq:qp_mat}, and solved the resulting \gls{miqp} using Gurobi solver~\cite{gurobi} on a 12th~Gen Intel Core i7-12700H with 32\,GB RAM; the average solve time of 1.81\,ms per step (solver-limited upper bound $\approx$ 552\,Hz). The simulation ran at 30\,Hz and the \gls{A} node at 50\,Hz.

\subsubsection{Sequence} Each mode begins with a fixed 120-second training session where the user can familiarize with the controls and freely move around. After training, participants begin a performance session where they are tasked to sequentially reach goals (green circles) in sequence as quickly as possible (shown in~\figref{fig:game_layout}), with only one goal appearing at a time.
\subsubsection{Collisions and Goals} A collision is triggered when the robot enters a black region, causing it to respawn at the initial or last goal position.  A goal is reached only when the robot fully enters the green circle at under $0.9$ units/s.

\subsubsection{Termination} A session ends either when all goals are reached (performance session) or when the time limit is reached (training session). For each performance session, we recorded \gls{cd} and collision count. After the session, the user completed the feedback survey. 

\subsection{User Study Setup}
Participants were divided into two groups: Group A experienced their first sequence (training + performance session) with \gls{NA}, followed by a second sequence with the \gls{A}, while Group B followed the reverse order. Each sequence began with a dedicated training session before the corresponding performance session.
\subsubsection{Participants}
The study included 66 participants, compensated with a free lunch. All surveys were optional. Most were aged 25–34 (54.5\%), followed by 18–24 (36.4\%) and 35–44 (9.1\%). The sample was predominantly male (71.2\%), with female participants comprising 28.8\%. Professionally, the majority were students (41.5\%), followed by engineers (30.8\%), researchers (16.9\%), and others (10.8\%). Regarding prior experience, 43.9\% of participants reported rarely playing video games (less than once every two months), while only 10.6\% played daily. Joystick use was even less common: 77.3\% of participants used a joystick rarely, and 3\% daily. The study was approved by our institution's research ethics committee, and informed consent was obtained from all participants.

\subsubsection{Survey}
Participants completed a survey after each simulation session, which included \gls{RTQ} listed in \gls{tlx}~\cite{hart1988development, hart2006nasa}, \gls{AQ}, and two \gls{OEQ} were asked to capture qualitative feedback. All items except the \gls{OEQ} were rated on a five-point Likert scale. For \gls{RTQ} ranging from ``Failure" to ``Perfect" for \gls{sc} and ``Very Low" to ``Very High" for the rest. For \gls{AQ}, from ``Strongly Disagree" to ``Strongly Agree". The exact questions asked are:
\glsreset{sc}

\begin{small}    
\begin{table}[H]
\centering
\setlength{\tabcolsep}{4pt}
\renewcommand{\arraystretch}{1.2}
\begin{tabular}{>{\centering\arraybackslash}m{0.6cm} m{11cm}}
\textbf{\gls{RTQ}} &
\begin{minipage}[t]{\linewidth}
\raggedright
\textbf{\gls{md}:} \textit{How mentally demanding was the task?} \\
\textbf{\gls{pd}:} \textit{How physically demanding was the task?} \\
\textbf{\gls{td}:} \textit{How rushed was the task pace?} \\
\textbf{\gls{sc}:} \textit{How successful were you?} \\
\textbf{\gls{eff}:} \textit{How hard did you work?} \\
\textbf{\gls{fr}:} \textit{How stressed or annoyed were you?}
\end{minipage} \\[0.2em]  

\textbf{\gls{AQ}} &
\begin{minipage}[t]{\linewidth}
\raggedright
\textbf{\gls{sf}: } \textit{I felt safe avoiding obstacles.} \\
\textbf{\gls{ct}: } \textit{The agent accurately followed my actions.}
\end{minipage} \\[0.2em]  

\textbf{\gls{OEQ}} &
\begin{minipage}[t]{\linewidth}
\raggedright
\textit{The assistive controller affected my performance by}~\dashuline{\textit{[text]}}. \\
\textit{My experience with and without assistive controller was}~\dashuline{\textit{[text]}}.
\end{minipage}
\end{tabular}
\vspace{-0.3cm}
\end{table}
\end{small}

\subsection{Evaluation and Testing Criteria}
\glsreset{cd}
To evaluate our hypotheses, we used survey responses (\gls{RTQ}, \gls{AQ}) for \hyp{1}–\hyp{3} and the metric \gls{cd} for \hyp{4}, supplemented by open-ended questions (\gls{OEQ}) for qualitative insights. We applied the \gls{WSRT} test for paired samples and the \gls{MWT} test for unpaired samples. Normality was assessed using the \gls{SWT} test.

Based on our predefined hypotheses, we conducted directional statistical tests to assess whether the median values under the \gls{A} were significantly higher or lower than those under the \gls{NA}. \tabref{tab:qual_pool_Survey} outlines the null hypotheses (\Hnullgeq~or~\Hnullleq) associated with each question and the corresponding hypothesis (\hyp{1}–\hyp{4}) it supports. A two-sided test (\Hnulleq) was applied only in cases where assessing distributional symmetry was relevant. The significance level for each statistical test is categorized as follows: $\alpha \leq$ \pone\ (significant), $\alpha \leq$ \ptwo\ (highly significant), and $\alpha \leq$ \pthree\ (very highly significant).

\section{User Study Results}\label{sec:results}

We evaluated the \gls{A}’s effectiveness through qualitative analysis for \hyp{1}–\hyp{3} and quantitative analysis for \hyp{4} to assess overall impact. 

\glsreset{WSRT}
\glsreset{MWT}

\paragraph{Qualitative Analysis}
To assess subjective experience and workload, we aggregated responses across all participants and analyzed the \gls{RTQ}, \gls{AQ}, and \gls{OEQ} surveys, targeting hypotheses \hyp{1}--\hyp{3}.

 As shown in \figref{fig:spider_agg}, \gls{A} consistently outperformed \gls{NA} across all questions supporting our hypothesis \hyp{1}--\hyp{3}.
\begin{table*}
\centering
\setlength{\tabcolsep}{4pt}
\begin{tabular}{l|cccccccccc}
 & \gls{md} & \gls{pd} & \gls{td} & \gls{sc} & \gls{eff} & \gls{fr} & \gls{sf} & \gls{ct} & \gls{cd}  \\
\midrule
Null Hypothesis (\Hnull)       & \Hnullgeq & \Hnullgeq &  \Hnullgeq & \Hnullleq & \Hnullgeq & \Hnullgeq & \Hnullleq & \Hnullleq & \Hnullgeq  \\ 
Alternate Hypothesis (\hyp{})       & \hyp{1}   & \hyp{1}   & \hyp{1}   & \hyp{1}   & \hyp{1}   & \hyp{1}   & \hyp{2}   & \hyp{3}   & \hyp{4}    \\
Significance ($\alpha$)     & \pthree   & \pthree   & \pone     & \pthree   & \pthree   & \pthree   & \pthree   & \pthree   & \pthree  \\
Sample Size ($\mathbf{N}$) & 66        & 65        & 64        & 60        & 60        & 61        & 62        & 66        & 66       \\ 
\bottomrule
\end{tabular}
\caption{
\gls{WSRT} test results comparing \gls{NA} vs. \gls{A}.}
\label{tab:qual_pool_Survey}
\figvspace{}
\end{table*}
\pgfplotstableread[col sep=comma]{./images/results/spider_plot/with_assistive.csv}\datatableWith
\pgfplotstableread[col sep=comma]{./images/results/spider_plot/without_assistive.csv}\datatableWithout
\definecolor{crimson2143940}{RGB}{214,39,40}
\definecolor{darkgray176}{RGB}{176,176,176}
\definecolor{darkorange25512714}{RGB}{255,127,14}
\definecolor{darkturquoise23190207}{RGB}{23,190,207}
\definecolor{forestgreen4416044}{RGB}{44,160,44}
\definecolor{goldenrod18818934}{RGB}{188,189,34}
\definecolor{gray127}{RGB}{127,127,127}
\definecolor{lightblue}{RGB}{173,216,230}
\definecolor{lightgray204}{RGB}{204,204,204}
\definecolor{mediumpurple148103189}{RGB}{148,103,189}
\definecolor{orchid227119194}{RGB}{227,119,194}
\definecolor{sienna1408675}{RGB}{140,86,75}
\definecolor{steelblue31119180}{RGB}{31,119,180}





\newcommand{\polarplotfromtable}[2]{%
\begin{tikzpicture}[scale = 0.48]
\definecolor{crimson2143940}{RGB}{214,39,40}
\definecolor{darkgray176}{RGB}{176,176,176}
\definecolor{darkorange25512714}{RGB}{255,127,14}
\definecolor{darkturquoise23190207}{RGB}{23,190,207}
\definecolor{forestgreen4416044}{RGB}{44,160,44}
\definecolor{goldenrod18818934}{RGB}{188,189,34}
\definecolor{gray127}{RGB}{127,127,127}
\definecolor{lightblue}{RGB}{173,216,230}
\definecolor{lightgray204}{RGB}{204,204,204}
\definecolor{mediumpurple148103189}{RGB}{148,103,189}
\definecolor{orchid227119194}{RGB}{227,119,194}
\definecolor{sienna1408675}{RGB}{140,86,75}
\definecolor{steelblue31119180}{RGB}{31,119,180}

\begin{polaraxis}[
legend cell align={left},
legend style={fill opacity=0.8, draw opacity=1, text opacity=1, at={(0.1,0.1)}, draw=darkgray},
unbounded coords=jump,
x grid style={black},
xmin=-40, xmax=405,
xtick style={color=black},
xtick={0,45,...,315},
xticklabel shift=-1.0pt,
xticklabels={
      \glsentryname{md}~$\downarrow$,
      \glsentryname{pd}~$\downarrow$,
      \glsentryname{td}~$\downarrow$,
      \glsentryname{sc}~$\uparrow$,
      \glsentryname{eff}~$\downarrow$,
      \glsentryname{fr}~$\downarrow$,
      \glsentryname{sf}~$\uparrow$,
      \glsentryname{ct}~$\uparrow$
    },
xticklabel style={
  font=\tiny,
  /pgf/number format/fixed,
  rotate=0,
  /pgfplots/every major tick label/.append style={
    /utils/exec={
        \pgfmathtruncatemacro{\tickint}{\tick}
        \ifnum\tickint=0 \pgfkeys{/tikz/anchor=west} \fi
        \ifnum\tickint=45 \pgfkeys{/tikz/anchor=west} \fi
        \ifnum\tickint=90 \pgfkeys{/tikz/anchor=west} \fi
        \ifnum\tickint=135 \pgfkeys{/tikz/anchor=south west} \fi
        \ifnum\tickint=180 \pgfkeys{/tikz/anchor=south} \fi
        \ifnum\tickint=225 \pgfkeys{/tikz/anchor=south east} \fi
        \ifnum\tickint=270 \pgfkeys{/tikz/anchor=east} \fi
        \ifnum\tickint=315 \pgfkeys{/tikz/anchor=north east} \fi
    }
  }
},
y grid style={darkgray},
ytick style={color=black},
ytick={1,2,3,4,5},
yticklabel style={
  font=\tiny,
  /pgf/number format/fixed,
  /utils/exec={
    \pgfmathtruncatemacro{\tickint}{\tick}
    \ifnum\tickint=1 \pgfkeys{/tikz/yshift=-1pt}\fi
    \ifnum\tickint=2 \pgfkeys{/tikz/yshift=-2pt}\fi
    \ifnum\tickint=3 \pgfkeys{/tikz/yshift=-3pt}\fi
    \ifnum\tickint=4 \pgfkeys{/tikz/yshift=-4pt}\fi
    \ifnum\tickint=5 \pgfkeys{/tikz/yshift=-5.0pt}\fi
  }
},
ymin=0.8,
ymax=6.0,
 axis line style={draw=none}, 
  legend columns=3,
  legend style={
    at={(0.5,-0.2)}, anchor=north, 
    /tikz/every even column/.append style={column sep=1em},
    draw=gray,
    fill=white,
    font=\small
  },
]

\addplot [thick, blue] table[
    x expr=\coordindex*45,
    y=Median
] {#1};

\addplot[name path=IQR_Upper, draw=none] table[
    x expr=\coordindex*45,
    y=IQR_Upper
] {#1};

\addplot[name path=IQR_Lower, draw=none] table[
    x expr=\coordindex*45,
    y=IQR_Lower
] {#1};

\addplot[
    fill=blue, opacity=0.2
] fill between[of=IQR_Upper and IQR_Lower];

\addplot[name path=Upper_Tail_95, draw=none] table[
    x expr=\coordindex*45,
    y=Upper_Tail_95
] {#1};

\addplot[name path=Lower_Tail_5, draw=none] table[
    x expr=\coordindex*45,
    y=Lower_Tail_5
] {#1};

\addplot[
    fill=lightblue, opacity=0.4
] fill between[of=Upper_Tail_95 and Lower_Tail_5];

\addplot[|-|, dashed, thick, domain=0:225, samples=100, opacity=0.5] ({x}, {5.5});
\node[rotate=22.5,  fill=white, text=black,  anchor=center, inner sep=0.08mm] at (axis cs:112.5, 5.5) {\scriptsize \gls{RTQ}};

\addplot[|-|, dashed, thick, domain=270:315, samples=100, opacity=0.5] ({x}, {5.5});
\node[rotate=22.5,  fill=white, text=black,  anchor=center, inner sep=0.08mm] at (axis cs:292.5, 5.5) {\scriptsize \gls{AQ}};

\end{polaraxis}
\end{tikzpicture}
}


\begin{figure}
\begin{minipage}{0.5\textwidth}
\centering
\begin{tikzpicture}[scale = 0.8]
\begin{axis}[
    hide axis,
    xmin=0, xmax=1,
    ymin=0, ymax=1,
    legend columns=3,
    legend style={
        draw=none,
        fill=none,
        font=\small,
        /tikz/every even column/.append style={column sep=1em},
        anchor=center,
        at={(0.5,-0.1)}
    }
]
\addlegendimage{line legend, thick, blue}
\addlegendentry{Median}
\addlegendimage{area legend, draw=blue, fill=blue, opacity=0.2}
\addlegendentry{IQR (25\% - 75\%)}
\addlegendimage{area legend, draw=lightblue, fill=lightblue, opacity=0.4}
\addlegendentry{Tails (5\% - 95\%)}
\end{axis}
\end{tikzpicture}
\end{minipage}
\\[-0.3em]
\begin{minipage}{0.22\textwidth}
    \centering
    \polarplotfromtable{\datatableWithout}{Without Assistance}
    \\[-0.5em]
    {\small (a) With \gls{NA}.}
\end{minipage}
\hspace{0.2cm}
\begin{minipage}{0.22\textwidth}
    \centering
    \polarplotfromtable{\datatableWith}{With Assistance}
    \\[-0.5em]
    {\small (b) With \gls{A}.}
\end{minipage}
\vspace{-0.35em}
\caption{\gls{RTQ} and \gls{AQ} responses: \gls{NA} vs. \gls{A}. Arrows show the desirable direction: $\uparrow$ = higher is better, $\downarrow$ = lower is better.}
\label{fig:spider_agg}
\figvspace{}
\end{figure}
\glsreset{td}
To further validate these results, a paired one-tailed \gls{WSRT} test confirmed statistically significant improvements for \gls{A} across all metrics (\tabref{tab:qual_pool_Survey}), with \vhs difference ($\alpha< $ \pthree), except for \gls{td}, which only showed \ms difference ($\alpha < $ \pone). 
In addition, most of the \gls{OEQ} responses from the users anecdotally reinforced that \gls{A} reduced the workload, made them feel much safer, and kept them feeling in control. One user reported: 

\begin{small}
\begin{quote}
\textit{``It is interesting with assistive system when the agent can avoid the obstacles by itself to make the player feel safe and secured despite how careless the player is. The agent was driven faster to reach the goal.''}
\end{quote}
\end{small}

\paragraph{Quantitative Analysis}
\glsreset{cd}
\glsreset{SWT}
To evaluate hypothesis \hyp{4}, we used \gls{OEQ} and recorded experimental data to calculate task \gls{cd} metric. For a fair comparison, \gls{cd} under the \gls{NA} was computed only from collision-free trajectories, excluding any collision or respawn delays. No collisions occurred with \gls{A}, supporting its safety guarantees.



\begin{figure}
\begin{minipage}{0.22\textwidth}
\centering


\begin{tikzpicture}[scale = 0.365]

\definecolor{darkgray176}{RGB}{176,176,176}
\definecolor{darkslategray63}{RGB}{63,63,63}
\definecolor{lightgray204}{RGB}{204,204,204}
\definecolor{peru22412844}{RGB}{224,128,44}
\definecolor{steelblue49115161}{RGB}{49,115,161}

\begin{axis}[
legend cell align={left},
legend style={
  font=\huge,
  fill opacity=0.8,
  draw opacity=1,
  text opacity=1,
  at={(0.55,0.97)},
  anchor=north west,
  draw=lightgray204
},
ymajorgrids=true,
y grid style={dashed, gray, opacity=0.8},
axis lines=left, 
tick align=outside,
tick pos=left,
tick label style={font=\huge},
x grid style={darkgray176},
xmin=-0.06, xmax=0.8,
width=8cm,
height=6.5cm,
scale only axis, 
trim axis left,
trim axis right,
xtick=\empty,
y grid style={darkgray176},
ylabel={Completion Duration (s)},
ylabel style={font=\LARGE},
ymin=70, ymax=500,
ytick style={color=black},
ytick distance=50,
]

\path [draw=darkslategray63, fill=peru22412844, semithick]
(axis cs:0,149.75)
--(axis cs:0.392,149.75)
--(axis cs:0.392,211.275)
--(axis cs:0,211.275)
--cycle;

\path [draw=darkslategray63, fill=steelblue49115161, semithick]
(axis cs:0.4,135.35)
--(axis cs:0.792,135.35)
--(axis cs:0.792,162.375)
--(axis cs:0.4,162.375)
--cycle;

\addlegendimage{line legend, thick, line width=5pt, draw=peru22412844}
\addlegendentry{\gls{NA}}

\addlegendimage{line legend, thick, line width=5pt, draw=steelblue49115161}
\addlegendentry{\gls{A}}

\addplot [semithick, darkslategray63, forget plot]
table {%
0.196 149.75
0.196 117.5
};
\addplot [semithick, darkslategray63, forget plot]
table {%
0.196 211.275
0.196 294.1
};
\addplot [semithick, darkslategray63, forget plot]
table {%
0.098 117.5
0.294 117.5
};
\addplot [semithick, darkslategray63, forget plot]
table {%
0.098 294.1
0.294 294.1
};
\addplot [black, mark=diamond*, mark size=2.5, mark options={solid,fill=darkslategray63}, only marks]
table {%
0.196 384.7
0.196 353.6
0.196 465
};
\addlegendentry{Outliers}

\addplot [semithick, darkslategray63, forget plot]
table {%
0.596 135.35
0.596 120.8
};
\addplot [semithick, darkslategray63, forget plot]
table {%
0.596 162.375
0.596 202.4
};
\addplot [semithick, darkslategray63, forget plot]
table {%
0.498 120.8
0.694 120.8
};
\addplot [semithick, darkslategray63, forget plot]
table {%
0.498 202.4
0.694 202.4
};
\addplot [black, mark=diamond*, mark size=2.5, mark options={solid,fill=darkslategray63}, only marks, forget plot]
table {%
0.596 211.2
0.596 205.5
0.596 228.8
0.596 249.4
0.596 207
0.596 208.2
};
\addplot [very thick, black, forget plot]
table {%
0 174.7
0.392 174.7
};
\addplot [very thick, black, forget plot]
table {%
0.4 142.9
0.792 142.9
};
\end{axis}
\end{tikzpicture}

\captionof{figure}{Aggregated Survey.}
\label{fig:pooled}
\end{minipage}
\hspace{0.21cm}
\begin{minipage}{0.22\textwidth}
\centering

\begin{tikzpicture}[scale = 0.365]
\definecolor{darkgray176}{RGB}{176,176,176}
\definecolor{darkslategray63}{RGB}{63,63,63}
\definecolor{lightgray204}{RGB}{204,204,204}
\definecolor{peru22412844}{RGB}{224,128,44}
\definecolor{steelblue49115161}{RGB}{49,115,161}

\begin{axis}[
legend cell align={left},
legend style={
  font=\huge,
  fill opacity=0.8,
  draw opacity=1,
  text opacity=1,
  at={(0.6,0.97)},
  anchor=north west,
  draw=lightgray204
},
ymajorgrids=true,
y grid style={dashed, gray, opacity=0.8},
axis lines=left, 
tick align=outside,
tick pos=left,
tick label style={font=\huge},
x grid style={darkgray176},
xmin=-0.1, xmax=1.8,
width=9cm,
height=6.5cm,
scale only axis, 
trim axis left,
trim axis right,
xtick=\empty,
y grid style={darkgray176},
 ylabel style={font=\huge},
ymin=70, ymax=500,
ytick style={color=black},
ytick distance=50,
]


\path [draw=darkslategray63, fill=peru22412844, semithick]
(axis cs:0,163.8)
--(axis cs:0.392,163.8)
--(axis cs:0.392,223.5)
--(axis cs:0,223.5)
--cycle;

\path [draw=darkslategray63, fill=peru22412844, semithick]
(axis cs:0.4,140.7)
--(axis cs:0.792,140.7)
--(axis cs:0.792,190.4)
--(axis cs:0.4,190.4)
--cycle;

\addlegendimage{line legend, thick, line width=5pt, draw=peru22412844}
\addlegendentry{\gls{NA}}
\addplot [semithick, darkslategray63, forget plot]
table {%
0.196 163.8
0.196 132.8
};
\addplot [semithick, darkslategray63, forget plot]
table {%
0.196 223.5
0.196 285.1
};
\addplot [semithick, darkslategray63, forget plot]
table {%
0.098 132.8
0.294 132.8
};
\addplot [semithick, darkslategray63, forget plot]
table {%
0.098 285.1
0.294 285.1
};
\addplot [black, mark=diamond*, mark size=2.5, mark options={solid,fill=darkslategray63}, only marks, forget plot]
table {%
0.196 384.7
0.196 353.6
};

\addplot [semithick, darkslategray63, forget plot]
table {%
0.596 140.7
0.596 117.5
};
\addplot [semithick, darkslategray63, forget plot]
table {%
0.596 190.4
0.596 239.5
};
\addplot [semithick, darkslategray63, forget plot]
table {%
0.498 117.5
0.694 117.5
};
\addplot [semithick, darkslategray63, forget plot]
table {%
0.498 239.5
0.694 239.5
};
\addplot [black, mark=diamond*, mark size=2.5, mark options={solid,fill=darkslategray63}, only marks, forget plot]
table {%
0.596 465
0.596 294.1
0.596 276.6
};

\addplot [very thick, black, forget plot]
table {%
0 192
0.392 192
};
\addplot [very thick, black, forget plot]
table {%
0.4 158.6
0.792 158.6
};

\path [draw=darkslategray63, fill=steelblue49115161, semithick]
(axis cs:1,137.1)
--(axis cs:1.392,137.1)
--(axis cs:1.392,156.4)
--(axis cs:1,156.4)
--cycle;

\path [draw=darkslategray63, fill=steelblue49115161, semithick]
(axis cs:1.4,135.3)
--(axis cs:1.792,135.3)
--(axis cs:1.792,162.6)
--(axis cs:1.4,162.6)
--cycle;

\addlegendimage{line legend, thick, line width=5pt, draw=steelblue49115161}
\addlegendentry{\gls{A}}

\addplot [semithick, darkslategray63, forget plot]
table {%
1.196 137.1
1.196 124.8
};
\addplot [semithick, darkslategray63, forget plot]
table {%
1.196 156.4
1.196 182.6
};
\addplot [semithick, darkslategray63, forget plot]
table {%
1.098 124.8
1.294 124.8
};
\addplot [semithick, darkslategray63, forget plot]
table {%
1.098 182.6
1.294 182.6
};
\addplot [black, mark=diamond*, mark size=2.5, mark options={solid,fill=darkslategray63}, only marks, forget plot]
table {%
1.196 189.7
1.196 194.3
1.196 202.4
1.196 205.5
1.196 228.8
1.196 207
};

\addplot [semithick, darkslategray63, forget plot]
table {%
1.596 135.3
1.596 120.8
};
\addplot [semithick, darkslategray63, forget plot]
table {%
1.596 162.6
1.596 182.8
};
\addplot [semithick, darkslategray63, forget plot]
table {%
1.498 120.8
1.694 120.8
};
\addplot [semithick, darkslategray63, forget plot]
table {%
1.498 182.8
1.694 182.8
};
\addplot [black, mark=diamond*, mark size=2.5, mark options={solid,fill=darkslategray63}, only marks]
table {%
1.596 211.2
1.596 249.4
1.596 208.2
};
\addlegendentry{Outliers}

\addplot [very thick, black, forget plot]
table {%
1 143
1.392 143
};
\addplot [very thick, black, forget plot]
table {%
1.4 142.8
1.792 142.8
};

\node[align=center, inner sep=0pt, font=\huge, rotate=0] at (axis cs:0.196, 95) {A};
\node[align=center, inner sep=0pt, font=\huge, rotate=0] at (axis cs:0.596, 95) {B};
\node[align=center, inner sep=0pt, font=\huge, rotate=0] at (axis cs:1.196, 95) {A};
\node[align=center, inner sep=0pt, font=\huge, rotate=0] at (axis cs:1.596, 95) {B};
\end{axis}
\end{tikzpicture}
\vspace{-0.4cm}
\captionof{figure}{Group-wise Survey.}
\label{fig:group}
\end{minipage}
\vspace{-0.2cm}
\figvspace{}
\end{figure}
This analysis revealed that \gls{A} significantly reduced \gls{cd}, supporting hypothesis \hyp{4}. As visualized in \figref{fig:pooled}, participants completed the task faster under \gls{A}. To validate these findings, we used paired data across all participants. The two-sided \gls{SWT} test yielded \vhs difference ($\alpha < $ \pthree), rejecting the null hypothesis (\Hnulleq) of normality, prompting the use of non-parametric one-tailed \gls{WSRT} test, which confirmed a \vhs reduction in \gls{cd} when using \gls{A} ($\alpha < $ \pthree). 
Feedback from the \gls{OEQ} further supports this trend. Many users felt safer with \gls{A}, which reduced perceived risk and made them less conservative, resulting in lower \gls{cd}. One user noted:

\begin{small}
\begin{quote}
\textit{``With no assistive system, I tend to choose more conservative route-move slower, always ready to stop and take detour if it can avoid collision. With assistive system, I tend to choose more aggressive (and shorter) route."}
\end{quote}
\end{small}

\paragraph{Learning effect}
To investigate how prior experience in one condition influences performance in the other, we conducted additional group-wise analyses using unpaired data within each group. Qualitative group-wise results were generally consistent with the paired data findings but did not reach significance for \gls{td}, likely due to reduced statistical power in the smaller subgroups. In contrast, quantitative group-wise results statistically mirrored the paired analysis, confirming that \gls{A} significantly reduced task \gls{cd} across both groups, as shown in \figref{fig:group}. Additionally, the analysis revealed a learning effect under \gls{NA}: participants who first used the \gls{A} took longer time when later operating without it, suggesting a potential reliance. No such effect was observed under \gls{A}—prior experience did not affect \gls{cd}—indicating that \gls{A} supports consistent behavior regardless of potential earlier experience without it.

\section{Real-World Experiment}
Shared control robotic applications such as surgery, industrial assembly, and hazardous material handling often require strict constraint satisfaction under disturbances. To demonstrate relevance in such settings, we evaluate our method on a Franka Panda manipulator (\figref{fig:real_setup}). Measurement noise, delay, and unmodeled dynamics are treated as bounded disturbances, with bounds found experimentally.
\begin{figure}[t]
    \centering
    \begin{subfigure}[t]{0.47\linewidth}
        \centering
        \raisebox{0.25\height}{\input{images/results/real_experiment/ws_setup}}
        \caption{}
        \label{fig:real_setup}
    \end{subfigure}%
    \hfill
    \begin{subfigure}[t]{0.49\linewidth}
        \centering
        \input{images/results/real_experiment/map_ws}
        \vspace{-1.2em}
        \caption{}
        \label{fig:real_traj}
    \end{subfigure}
    \caption{Real experiment setup. (a) shows the maze workspace, and (b) the trajectory with disturbance bound (red circle).}
\figvspace{}
\end{figure}

The user was tasked to navigate the end-effector, fitted with a $19\mathrm{mm}$ peg, through a maze of three rectangular blocks with $32\mathrm{mm}$ passages, leaving $6.5\mathrm{mm}$ clearance per side, reduced to about $3.5\mathrm{mm}$ under $\pm 3\mathrm{mm}$ disturbance bounds. Five trials were conducted, and the trajectory from one trial is shown in~\figref{fig:real_traj}. Velocity is constrained to $[-0.05,0.05]\mathrm{m/s}$ (disturbance $\pm 0.01\mathrm{m/s}$) and acceleration to $[-1,1]\mathrm{m/s^2}.$ A Cartesian controller drives the end-effector in the $X$–$Y$ plane using assistive inputs $u^a_k$ generated by the proposed \gls{A} framework, with states $(X,Y,v^x,v^y)$ governed by~\eqref{eq:agent_dyna}. Acceleration-level control is used instead of position-level control, making the task more challenging.
\begin{figure}
    \centering
\input{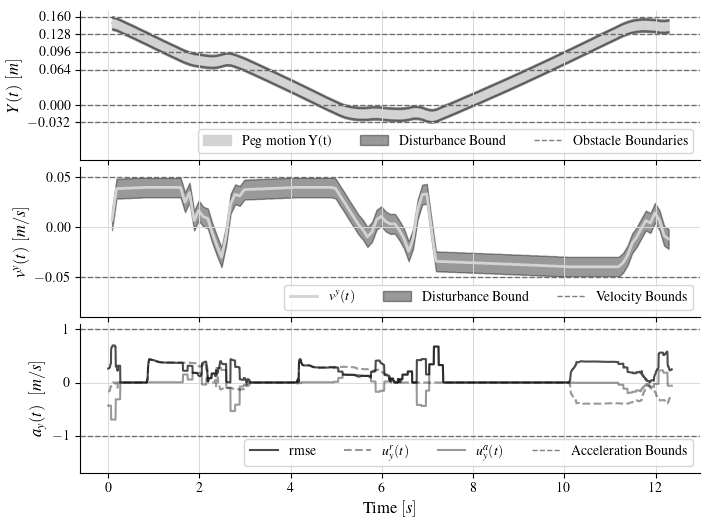}
            \caption{$Y$-axis position, velocity, and control inputs (top–bottom), with RMSE of user vs. controller input.}
    \label{fig:fig_y_axis}
    \figvspace{}
\end{figure} 
To analyze the controller’s behavior, we focus on the $y$-axis, where the manipulator covers a larger distance. As the $x$- and $y$-dynamics are decoupled, it is sufficient to examine this axis. \figref{fig:fig_y_axis} shows the evolution of position, velocity, and control inputs(top-bottom).

The controller’s behavior can be categorized into three regions. In region $(P)$, the controller intervenes to avoid wall collisions in narrow passages. In region $(V)$, the controller enforces velocity limits whenever user inputs $u^r_y$ would exceed the limits. In region $(\varnothing)$, the controller followed the user inputs exactly (RMSE $\approx 0$), as the end-effector was well aligned with the passage and required no correction to satisfy the limits.  

The experiment confirms that the controller enforces safety and constraint satisfaction with minimal intervention under disturbance.

Importantly, the system remains safe under arbitrary user inputs, as the controller intervenes only when necessary. This property is also relevant in teleoperation with communication delays, where deploying the controller locally ensures safety.
\section{Conclusions}\label{sec:conclusions}
We proposed a constraint-aware \gls{A} framework for real-time shared control with formal safety guarantees under structured non-convex constraints. 

{The user study confirmed the effectiveness of the \gls{A}. Qualitative results showed improved safety, control, and reduced workload, while quantitative analysis demonstrated faster task completion with zero collisions. Overall, the \gls{A} enabled more direct routes without loss of perceived control, reflecting greater trust and confidence.} 

The real-world manipulator experiment further confirmed the framework’s practicality by showing that it can maintain safety under disturbances and operate reliably with structured non-convex constraints. This highlights its applicability to assistive tasks and teleoperation scenarios where safety assurance is paramount.

While the approach assumes known linear time-invariant dynamics, it establishes a principled foundation for constraint-aware shared control. Thus, extending the work to dynamic and uncertain settings, nonlinear models, and learning-based safety sets is an appealing avenue for the future, towards transparent, reliable, and minimally intrusive real-time assistance.

\bibliographystyle{IEEEtran}
\bibliography{IEEEabrv,Bibliography.bib}

\end{document}